\pdfoutput=1
\documentclass[11pt]{article}

\usepackage[]{acl}

\usepackage{times}
\usepackage{latexsym}

\usepackage[T1]{fontenc}
\usepackage[utf8]{inputenc}

\usepackage{ulem}
\usepackage{times}
\usepackage{latexsym}
\usepackage{url}

\usepackage{graphicx}
\graphicspath{ {./images/} }
\usepackage[table,xcdraw]{}
\usepackage{microtype}
\usepackage{float}

\usepackage{booktabs}
\usepackage{multirow} %合并多行单元格的宏包
\usepackage{makecell}
\usepackage{tabularx}
\usepackage{threeparttable}
\usepackage{amssymb}
\usepackage{marvosym}
\usepackage{subfigure}
\usepackage{soul}

\title{Evaluating Discourse Cohesion in Pre-trained Language Models}

\author{Jie He\footnotemark[2]  , Wanqiu Long\footnotemark[2]  , \and Deyi Xiong\footnotemark[3] \  \\
  \footnotemark[2]\ \  University of Edinburgh, Edinburgh, UK\\
  \footnotemark[3]\ \  College of Intelligence and Computing, Tianjin University, Tianjin, China\\
  \ \tt {j.he@ed.ac.uk, Wanqiu.long@ed.ac.uk, dyxiong@tju.edu.cn}
  }

\begin{document}
\maketitle
\begin{abstract}
Large pre-trained neural models have achieved remarkable success in natural language process (NLP), inspiring a growing body of research analyzing their ability from different aspects. In this paper, we propose a test suite to evaluate the cohesive ability of pre-trained language models. The test suite contains multiple cohesion phenomena between adjacent and non-adjacent sentences. We try to compare different pre-trained language models on these phenomena and analyze the experimental results, hoping more attention can be given to discourse cohesion in the future. 

\end{abstract}


% \keywords{natural language processing, datasets, discourse cohesion}

\maketitle
\section{Introduction}

Pre-trained language models have achieved remarkable success in many downstream tasks, including question answering \cite{etal-2019-multi}, reading comprehension \cite{etal-2019-enhancing-pre}, and machine translation \cite{sumita-2019-recycling}, inspiring a growing body of research analyzing their ability from different aspects \cite{2019-contextual,etal-2019-bert}. However, to our best knowledge, there is no existing work to evaluate whether the abilities of these models to identify and generate discourse cohesion. 

Cohesion is the foundation of an essay and an important form of showing style and character, and it is a semantic property of a document that represents the degree to which discourse entities are knit throughout the document \cite{Li2013TheAA,long-etal-2020-shallow,long-etal-2020-ted,https://doi.org/10.48550/arxiv.2205.07532}.
\citet{1976cohesion} defined cohesion as “the set of possibilities that exist in the language for making text hang together”. Cohesion occurs where the interpretation of some element in the discourse is dependent on that of another. For example, an understanding of the reference of a pronoun (he, she, it, etc.) requires to look back to something that has been said before. Through this cohesion relation, two text clauses or sentences are linked together. Therefore, cohesion plays an important role in discourse.  

However, to our best knowledge, existing available resources either only provide annotations for one cohesive phenomenon or mainly focus on lexical cohesion. For example,  \citet{Bos2011AnAC} annotate verbal phrase ellipsis; \citet{Martnez2016AnnotationOL} annotate lexical cohesion for both German and English texts. However, neither single cohesion phenomena nor just lexical cohesion can fully interpret the ability of models from the perspective of cohesion. 
 
 %Existing fusion datasets are small, which is perhaps why only few works have explored the application of supervised models to sentence fusion (Elsner and Santhanam, 2011; Thadani and
%McKeown, 2013).

Considering the above, this work has the following contributions:

\begin{itemize}
\setlength{\itemsep}{0pt}
\setlength{\parsep}{0pt}
\setlength{\parskip}{0pt}
\item We study discourse cohesion for pre-trained language models, which has been under-studied in previous works on representation learning, but is critical to language understanding and generation.

\item  We propose a test suite of cohesion including both grammatical and lexical cohesion phenomena. 

\item  We conduct a qualitative analysis of different pre-trained language models for their ability for multiple cohesion phenomena from both adjacent and non-adjacent sentences.

\end{itemize}

\begin{table*}
\renewcommand\arraystretch{1.25}
\setlength{\belowcaptionskip}{-7pt}

\centering
%\small
\footnotesize
% \scriptsize
\begin{tabular}{p{50pt}lp{300pt}c}
\bottomrule
cohesion phenomenon&Category&Example&Size\\
\hline
Repetition&adj&he decided to buy a  {{\textbf{\uline{pair}}}} of khakis. the  {{\textbf{\uline{pair}}}} he bought fit him perfectly. &200\\
&non-adj&Jude was very excited about his college graduation {{\textbf{\uline{ceremony}}}}. On the way to the arena, he got stuck in traffic. He only had an hour before the {{\textbf{\uline{ceremony}}}} started.&73\\
\hline
Synonyms&adj&jill became very  {{\textbf{\uline{scared}}}}. liam could tell jill was truly  {{\textbf{\uline{frightened}}}}.&200\\
&non-adj&She decided not to pursue the {{\textbf{\uline{matter}}}} and just keep the service. It was after all only \$12. But the {{\textbf{\uline{issue}}}} kept bothering her.&64\\
\hline
Ellipsis&adj&But {{\textbf{\uline{we}}}} have an interest in     {{\textbf{\uline{hiring him}}}}; I just don't know  {{\textbf{\uline{when}}}}. &200\\
&non-adj&Shawn felt that he could learn {{\textbf{\uline{to make the website on his own}}}}. Due to budget he could not pay a web designer. He took many web development classes to learn {{\textbf{\uline{how}}}}.&50\\
\hline
Substitution&adj& She wanted those  {{\textbf{\uline{cookies}}}}. She then decided to take {{\textbf{\uline{one}}}}.&200 \\  
&non-adj&She began to drink a few {{\textbf{\uline{beers}}}}.  He had never been a drinker. She encouraged him to drink {{\textbf{\uline{one}}}}.&61\\
\hline
Reference&adj&At first he did not like the {{\textbf{\uline{classes}}}}. however, over time he began to like {{\textbf{\uline{them}}}} a lot.&200\\
&non-adj&Once there {{\textbf{\uline{Jill}}}}  marveled at all the beauty. It was dangerous, but exciting. {{\textbf{\uline{She}}}} had a wonderful time on her trip to the Amazon.&51\\
\hline
Conjunction&adj&it was also cash only.  {{\textbf{\uline{therefore}}}} i had to turn around and go home.&200\\
&non-adj&The couple rented a yurt. It was very small. They did not like being so close. They left the Yurt. They rented a hotel {{\textbf{\uline{instead}}}}.&55\\

\hline
\toprule
\end{tabular}
\caption
{Examples of cohesion phenomena adopted in our test suite.  Repetition
and synonyms are lexical cohesion. Non-adj means the cohesion phenomenon is annotated between non-adjacent sentences, while adj refers to cohesion between adjacent sentences.}
\label{table-exam}
\end{table*}

\section{Related work} 

\noindent
\textbf{Discourse Cohesion Modeling}

\noindent
Some discourse cohesion phenomena have been applied in various NLP tasks. A thorough survey of related work on this is far beyond the scope of this paper. To name just a few,  \citet{etal-2019-good} study repetition and ellipsis in machine translation; \citet{etal-2019-discofuse} tried to bring the connection between two sentences closer by combining rule-based methods with coreference and conjunction. Similarly, there are also some works dedicated to the study of discourse phenomena. For example,  \citet{uryupina_artstein_bristot_cavicchio_delogu_rodriguez_poesio_2020}  annotated a broad range of anaphoric phenomena in a variety of genres. \citet{pishdad-etal-2020-coherent} studied the phenomenon of coherence at both the lexical and document levels. \citet{long2024leveraginghierarchicalprototypesverbalizer} focus on the multi-label problem in discourse analysis.  We are the first work to evaluate the performance of the pre-trained language model about multiple discourse cohesion phenomena.

\noindent
\textbf{Analysis towards Pre-trained Language Models}

\noindent
The boom of pre-trained language models has stimulated plenty of work to probe into the internal working mechanisms and capacities of pre-trained language models \cite{2019roberta,etal-2019-bert,lewis-etal-2020-bart,long-webber-2022-facilitating,long-etal-2024-multi,he2025mintqamultihopquestionanswering}. For example, \citet{jawr-etal-2019-bert} investigate the ability of these pre-trained models from the structure of language; \citet{etal-2019-linguistic,wart2019blimp} analyze those models from syntactic phenomena. \citet{etal-2019-evaluation} study whether sentence representations from pretrained language models contain contextual information.
%节省空间
Meanwhile, \citet{etal-2019-probing} test pre-trained language models for functional words within sentences.

However, although there are resources annotated for individual phenomena separately, there are not so many
annotated for several types of devices, so no existing work tries to simultaneously evaluate whether the pre-tained language models are good enough for identifying and generating different multiple cohesion phenomena and to  compare and analyze the results.

%In addition,  compared with the task of discourse evaluation %from another work  \citep{journals/corr/abs-1907-08672},  which %just evaluate the performance of pre-trained language models on %multiple discourse tasks, we attempt to explore more specific %components of discourse.

% In recent years, discourse has been extensively studied by RST\citep{Towardafunctionaltheoryoftextorganization} and PDTB\citep{2009recognizing} theories. These two theories mainly analyze the relationship between sentences. The hierarchical sentence structure constructed by RST can be used in the field of text summarization\citep{2019discourseaware}, while for pdtb Many studies have been devoted to better letting the model predict the relationship between pairs of sentences\citep{etal-2019-topic}. And some of the work is to use the discourse relationship between sentences to improve the pre-trained language model\citep{2020pretraining}. The translation of the discourse phenomenon for machine translation has also been explored by people\citep{voi-etal-2018-context}. What is similar to our work is Regarding the task of sentence fusion\citep{etal-2019-discofuse}, the relationship between the sentences is used to add connecting words to the two sentences to merge based on the rules. But we have studied more cohesion phenomena by comparison. The task of discourse-eval\citep{journals/corr/abs-1907-08672} is to evaluate the performance of pre-trained language models related to multiple discourse tasks, and we have explored the more specific components of discourse.

\section{Our Test Suite and its Annotation}

\subsection{Introduction} 

\citet{1976cohesion} describe five main types of cohesion in English,  which we adopt for our suite: reference, substitution, ellipsis, conjunction and lexical cohesion. Table 1 demonstrates the examples and size for the six cohesion phenomena covered in our test suite. The test suite contains 1554 cohesion examples in total. While cohesive cohesion have in principle noting to do with sentence boundaries \cite{1976cohesion}, we take into account cohesive relations between adjacent sentences/clauses as well as those between non-adjacent sentences. However, due to the data sparsity, there are 354 instances in total between non-adjacent sentences, while each phenomenon has 200 instances between adjacent sentences.

The cohesion examples for six cohesion phenomena in this test suite were all drawn from the ROC stories corpus \cite{etal-2016-corpus}. There are 50k five-sentence commonsense stories in this corpus. This corpus is a high quality collection of everyday life stories, which captures a rich set of  relations between daily events.

\subsection{Lexical Cohesion}

Lexical cohesion arises from the semantic relationship between words, as the chains of related words can generate the continuity of lexical meaning. Two typical ways of achieving this kind of cohesion is repetition and synonyms.

\noindent
\textbf{Repetition}: Repetition means the repeating of certain words or phrases. The task is to study the relationship between repeated words from two sentences, while our dataset for this phenomenon is on the nouns repetition.

\noindent
\textbf{Synonyms}: As for synonyms, it means there are related words that having the same connotations, implications, or reference in two sentences. Therefore, the task is to observe whether the synonyms from two sentences are magnets for each other in the models. In our test suite, the sentence pairs for this phenomenon include nouns indicating synonyms.

\subsection{Grammatical Coehsion}

Our grammatical cohesion tasks investigate whether the models have the ability to identify the anaphoric relationship between entities or how the sentences are connected with each other.

\noindent
\textbf{Reference}: Reference is a relationship between objects in which one object designates, or acts as a means by which to connect to or link to, another object.

\noindent
\textbf{Substitution}: Substitution generally occurs when one item within a text or discourse is replaced by another. The examples for this phenomenon are mainly represented by the substitution of nouns by using ``one''. For instance, ``this house is old. I will buy a new one''. 

\noindent
\textbf{Ellipis}: Ellipsis means the omission of one or more words that are obviously understood but that must be supplied to make a construction grammatically complete. For this part of the data, we use the sluice ellipsis dataset \cite{anand-mccloskey-2015-annotating}, which studies the omission after wh-words. 

\noindent
\textbf{Conjunction}: Unlike other grammatical cohesion phenomena, conjunction expresses a logical semantic relationship between two sentences rather than between words or structures. According to \citet{1976cohesion}, conjunction can be divided into 4 categories: additive, adversative, causal, and temporal. In our test set, we covered these 4 categories.

\noindent
\textbf{Markers}: Although without discourse markers, the meaning of the sentences would not be affected, they enable the connection between sentences to stick together. 

\begin{table*}

\centering
\footnotesize 
% \scriptsize
\begin{tabular}{l|p{11pt}p{28pt}p{11pt}p{28pt}|p{11pt}p{28pt}p{11pt}p{28pt}p{11pt}p{28pt}p{11pt}p{28pt}}
\bottomrule
&\multicolumn{2}{c}{Repetition}&\multicolumn{2}{c}{Synonym}&\multicolumn{2}{c}{Reference}&\multicolumn{2}{c}{Substitution}&\multicolumn{2}{c}{ellipsis}&\multicolumn{2}{c}{conjunction} \\
\textbf{Model}&adj&non-adj&adj&non-adj&adj&non-adj&adj&non-adj&adj&non-adj&adj&non-adj\\
\hline

BERT-base&0.690&0.493&0.240&0.391&0.830&0.510&0.365&0.262&0.421&0.180&0.235&0.364\\
BERT-large&0.730&0.644&0.270&0.469&0.850&0.608&0.470&0.328&0.455&0.280&0.340&0.455\\
BART-base&0.725&0.795&0.215&0.422&0.675&0.490&0.375&0.180&0.302&0.34&0.135&0.018\\
BART-large&0.710&0.740&0.250&0.500&0.715&0.627&0.390&0.230&0.302&0.260&0.100&0.145 \\
% Xlnet-base&0.940&0.821&0.649&0.542&0.727&0.523&0.617\\
% Xlnet-large&0.967&0.834&0.723&0.676&0.875&0.712&0.677\\
RoBERTa-base&0.780&0.712&0.325&0.469&0.790&0.804&0.545&0.377&0.624&0.540&0.395&\textbf{0.673}\\
RoBERTa-large&\textbf{0.815}&\textbf{0.836}&\textbf{0.430}&\textbf{0.594}&\textbf{0.855}&\textbf{0.863}&\textbf{0.665}&\textbf{0.393}&\textbf{0.678}&\textbf{0.600}&\textbf{0.485}&0.655\\
\hline
HUMAN&0.86&0.72&0.83&0.915&0.952&0.810&0.876&0.780&0.865&0.820&0.925&0.840 \\
\toprule
\end{tabular}
\caption
{Accuracy of the masked-word-prediction }
\label{table-dis}
\end{table*}

\subsection{Annotation}%标注

To construct the test suite, we hired 2 fluent English speakers to manually annotate data.

Since cohesion is something available in the surface structure, it is relatively easy to identify. Therefore, we were able to filter a great number of sentences without cohesion by using the  “cohesive devices”  and WordNet \cite{Fellbaum2000WordNetA}. Cohesive devices are words or phrases used to connect ideas between different parts of text. From Table 1, we can see ``one'', ``when'', ``how'', ``therefore'', etc. as  “cohesive devices”. WordNet was used to identify synonyms. 

However, the automatic filtering is just the first step. Human annotation is necessary since most automatically selected sentences have no cohesion. Before manual annotation, our annotation guidance and requirements were explained in detail to the annotators:
\begin{itemize}
\setlength{\itemsep}{0pt}
\setlength{\parsep}{0pt}
\setlength{\parskip}{0pt}

    \item The annotators are required to observe whether the sentence has corresponding phenomena. For example, the repetition phenomenon requires the nouns that refer to the same thing to appear twice in the sentence. The phenomenon of ellipsis requires ellipsis hint words (wh-words here) to appear in the sentence. 
    \item After identifying whether certain cohesion phenomenon is shown, the annotators needs to mark the two elements that convey cohesion. If the two elements that convey cohesion cannot be marked, the sentence would not be used.
\end{itemize}

To ensure annotation consistency, we compute the Kappa value and agreement rate between two annotators for agreement study. Before annotation, we randomly selected 500 examples as samples for pre-annotation, then two annotators labelled the text in terms of our annotation guidelines respectively. Finally, we got the average IAA and Cohen's kappa value for the two annotators' annotation, which is 91.3\% and 80.6\%.

% For examples, the annotators were required to guarantee no cohesion exists between adjacent sentences if what they annotate is the cohesive relation between non-adjacent sentences. While annotating, the annotators not only annotated the type of cohesion, but also mark the two elements that convey cohesion. In this way, references could be provided for our evaluation task.

% \begin{figure}
% \setlength{\belowcaptionskip}{-0.5cm} 
% \centering

% \label{fig1}
% \centering
% \includegraphics[width=7cm]{images/figure1.pdf}
% %\caption{fig1}

% \centering
% \caption{RoBERTa large performance on non-adjacent sentence.}
% \end{figure}

% \begin{figure}
% \setlength{\belowcaptionskip}{-0.5cm} 
% \centering

% \label{fig2}
% \centering
% \includegraphics[width=7cm]{images/figure2.pdf}
% %\caption{fig1}

% \centering
% \caption{RoBERTa large performance on adjacent sentence.}
% \end{figure}

\section{Experiments}

\subsection{Models}

We chose the pre-trained language model BERT \cite{den-etal-2019-bert}, BART \cite{lewis-etal-2020-bart} and RoBERTa \cite{2019roberta} as our evaluation models. The pretraining task of BART involves randomly shuffling the order of the original sentences and a novel in-filling scheme, where spans of text are replaced with a single mask token. While BERT and RoBERTa mainly differ in their training set size, BERT and BART is different in their training methods and model architectures.

\subsection{Cohesion Evaluation}
We would like to investigate whether the pretrained language models capture enough knowledge related to cohesion. We evaluated model performance via the prediction of masked words. A masked-word-prediction head (either fine-tuned or not) produces a probability distribution over its whole vocabulary via a softmax layer. We consider hit@1, namely the word filled with the highest probability when evaluating. If the hit@1 generated is able to link two clauses or sentences together, we think the model show the ability of identifying and generating cohesion. For example, in this example, "he decided to buy a pair of khakis. The [MASK] he bought fit him perfectly." ,  "pair"  would be expected to be filled when considering repetition.
 
Besides, to investigate whether the models utilize the context, we compare the probability of generating the target word with and without the previous sentences/clauses on the sub-testset of cohesion between adjacent sentences. In the example, "he decided to buy a pair of khakis. The [MASK] he bought fit him perfectly.", we compare the probability of generating the target word "pair"  with and without the span of "he decided to buy a pair of khakis". Finally, we got average probability of the target words for the six cohesion phenomena in both situations.

% \begin{figure}
% \setlength{\belowcaptionskip}{-0.7cm} 
% \centering

% \label{fig1}
% \centering
% \includegraphics[width=6cm]{images/figure1.pdf}
% %\caption{fig1}

% \centering
% \caption{RoBERTa large performance on non-adjacent sentence.}
% \end{figure}

% \begin{figure*}[p]
% \setlength{\belowcaptionskip}{-0.3cm}
% \centering
% \subfigure[]{
% \begin{minipage}[t]{0.65\linewidth}
% \setlength{\belowcaptionskip}{-0.5cm}
% \includegraphics[width=7cm]{images/figure1.pdf}
% \setlength{\abovecaptionskip}{-12pt} 
% %\caption{(a)}
% \label{figure-1}
% \end{minipage}%
% }%
% \subfigure[]{
% \begin{minipage}[t]{0.3\linewidth}
% \setlength{\belowcaptionskip}{-0.5cm}
% \includegraphics[width=7cm]{images/figure2.pdf}
% \setlength{\abovecaptionskip}{-12pt} 
% %\caption{(b)}
% \label{figure-2}
% \end{minipage}%
% }%
% \flushright
% \setlength{\abovecaptionskip}{-4pt} 
% \caption{Attention heatmaps for 7 types of discourse phenomena. }
% \end{figure*}

\subsection{Results}
Table 2 displays the result of our evaluation task. Firstly, we can see that RoBERTa is the best model in terms of their performance on all  cohesion phenomena. BART is inferior to BERT in many phenomena such as synonyms, reference, subsitution, ellipsis. This indicates that the pre-training task of BART may not be very helpful for understanding discourse cohesion phenomena. 
% \begin{figure}
% \setlength{\belowcaptionskip}{-0.5cm} 
% \centering

% \label{fig1}
% \centering
% \includegraphics[width=7cm]{images/figure1.pdf}
% %\caption{fig1}

% \centering
% \caption{RoBERTa large performance on non-adjacent sentence.}
% \end{figure}

% \begin{figure}
% \setlength{\belowcaptionskip}{-0.5cm} 
% \centering

% \label{fig2}
% \centering
% \includegraphics[width=7cm]{images/figure2.pdf}
% %\caption{fig1}

% \centering
% \caption{RoBERTa large performance on adjacent sentence.}
% \end{figure}

From table 2, we can see that conjunction, substitution, synonym and ellipsis are more complicated cohesion types, because the pre-trained language models are not good at them, compared with other cohesion phenomena. With regard to synonyms, it requires that the models not only can identify the cohesion but also have awareness of paraphrasing, which makes it difficult for the models. Looking at the data, we found that the RoBERTa tends to repeat the same word instead of generating another similar word to express the same meaning, even when it notices there is cohesion between the word that should be covered and the corresponding word. In other words, if the models fail to find other cohesive ways, they would try to repeat the words they identify to convey cohesion.

% \begin{figure}
% \setlength{\belowcaptionskip}{-0.7cm} 
% \centering

% \label{fig2}
% \centering
% \includegraphics[width=6cm]{images/figure2.pdf}
% %\caption{fig1}

% \centering
% \caption{RoBERTa large performance on adjacent sentence.}
% \end{figure}

Moreover, model performance on cohesion phenomena between adjacent sentences and non-adjacent sentences can be compared by looking at the Table 2. The models perform better for the cohesion phenomena between non-adjacent sentences instead of adjacent sentences except for substitution. It might be because additional sentences between the two cohesive elements provide context for the models to identify those cohesion phenomena. 

% In addition, as shown in Figure 2, if they can identify the cohesion, the models are less likely to directly use repetition for non-adjacent cohesion. The possible reason is that long distance makes it less likely to use repetition, with less impact from the original word.

\begin{table*}[tbp]
\centering
\scriptsize 
\begin{tabular}{l|ll|ll|ll|ll|ll|ll}
\bottomrule
&\multicolumn{2}{c|}{Repetition}&\multicolumn{2}{c|}{Synonym}&\multicolumn{2}{c|}{Reference}&\multicolumn{2}{c|}{Substitution}&\multicolumn{2}{c|}{ellipsis}&\multicolumn{2}{c}{conjunction} \\
\textbf{Model}&w/o-C&w/-C&w/o-C&w/-C&w/o-C&w/-C&w/o-C&w/-C&w/o-C&W-C&w/o-C&w/-C\\
\hline

BERT-base&0.085&0.510&0.083&0.173&\textbf{0.262}&0.664&0.061&0.266&0.257&0.338&0.050&0.082\\
BERT-large&0.116&0.557&0.100&0.209&0.238&\textbf{0.737}&0.060&0.363&\textbf{0.260}&0.399&\textbf{0.061}&\textbf{0.098} \\
BART-base&0.047&0.392&0.050&0.105&0.052&0.279&0.023&0.172&0.103&0.207&0.002&0.003\\
BART-large&0.045&0.309&0.061&0.128&0.067&0.337&0.031&0.209&0.127&0.233&0.002&0.003 \\
% Xlnet-base&0.940&0.821&0.649&0.542&0.727&0.523&0.617\\
% Xlnet-large&0.967&0.834&0.723&0.676&0.875&0.712&0.677\\
RoBERTa-base&0.109&0.585&0.106&0.223&0.155&0.507&0.062&0.407&0.221&0.457&0.009&0.031\\
RoBERTa-large&\textbf{0.144}&\textbf{0.662}&\textbf{0.114}&\textbf{0.268}&0.175&0.652&\textbf{0.079}&\textbf{0.515}&0.257&0.52&0.01&0.075\\

\toprule
\end{tabular}
\caption
{Probability of the target word with and without prior context.}
\label{table-dis1}
\end{table*}
\begin{figure*}[!htbp]
\setlength{\belowcaptionskip}{-0.3cm}
\centering
\subfigure[]{
\begin{minipage}[t]{0.65\linewidth}
\setlength{\belowcaptionskip}{-0.5cm}
\includegraphics[scale=0.35]{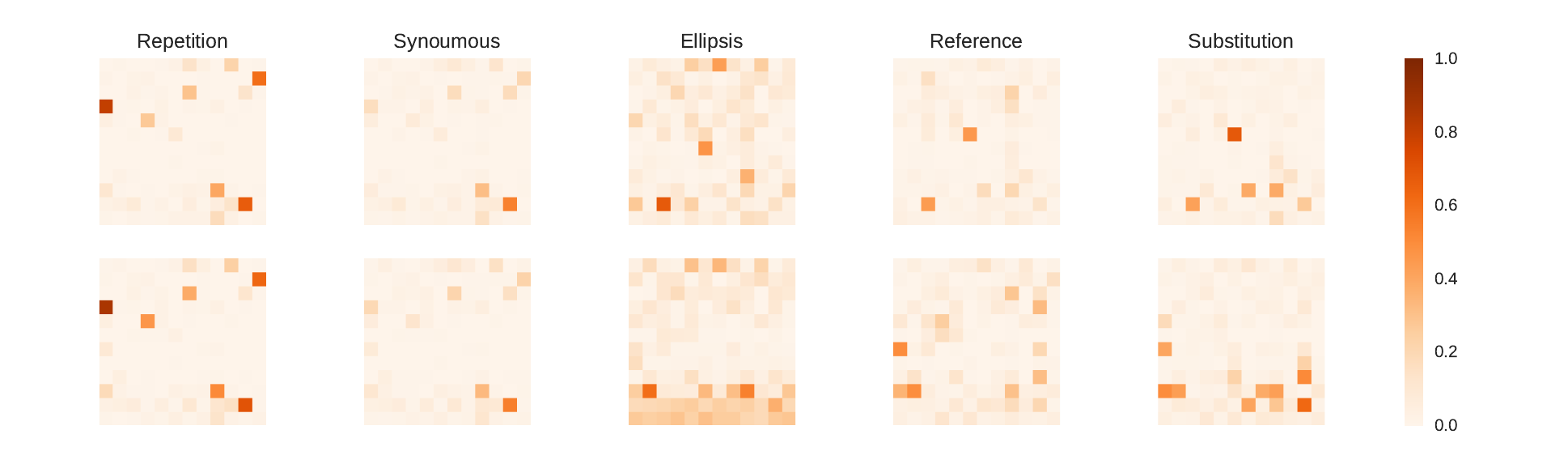}
\setlength{\abovecaptionskip}{-12pt} 
%\caption{(a)}
\label{first_5}
\end{minipage}%
}%
\subfigure[]{
\begin{minipage}[t]{0.2\linewidth}
\setlength{\belowcaptionskip}{-0.5cm}
\includegraphics[scale=0.35]{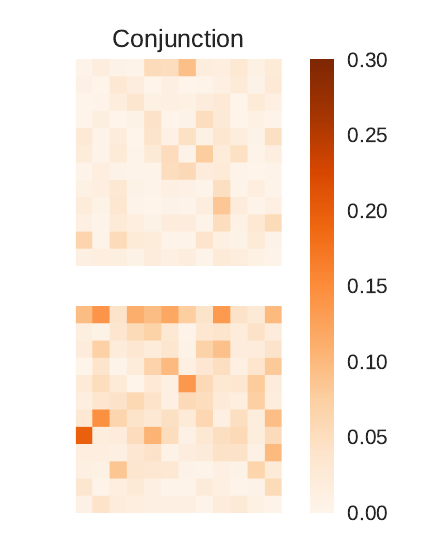}
\setlength{\abovecaptionskip}{-12pt} 
%\caption{(b)}
\label{later_2}
\end{minipage}%
}%
\flushright
\setlength{\abovecaptionskip}{-4pt} 
\caption{Attention heatmaps for 7 types of discourse phenomena. }
\end{figure*}
\section{The probability of generating the target word }

Table 3 gives us the information about the probability of generating the target word with and without providing the previous sentences/clauses. From the results of table 3, we can see without the previous sentence/clause, the possibilities of generating the target word for all cohesion phenomena are greatly decreased. Therefore,  there is strong cohesion between the target word in the second sentence and the corresponding word in the first sentence. However, the context provided by the first sentence have little positive impacts on BART for these cohesion phenomena, compared with other models.  

\section{Internal Analysis of BERT for Cohesion Phenomena}

% \subsection{Internal Analysis of BERT for Cohesion Phenomena}

For these 7 kinds of cohesion phenomena, we got some fine-grained information from the attention heatmap. The upper part of Figure \ref{first_5} indicates the attention between the words of  sentence/clause one and the words of the second sentence/clause two, while the below of Figure \ref{first_5} demonstrates the attention between the words of sentence two and sentence one. We note that repetition and synonym have great attention in both directions, with almost equivalent attention. This explains why the models are better at identifying these two cohesion phenomena. What's more, the attention mainly gather on the deeper layers, which might reflect the deeper layers of BERT capture more complex semantic features. 

In Figure \ref{later_2}, the upper part represents the attention between the first sentence and the conjunction word/discourse marker, whereas the below represents the attention between the second sentence and the conjunction word or discourse marker. The attention heatmap shows that much more attention can be seen between sentence two and the words, which means that the conjunction word or discourse marker is more closely related to the second sentence. However, it can be observed that the maximum attention of all head value for these two phenomena does not exceed 0.3, thus illustrating the poor performance of the pre-trained language models on these two phenomena is largely due to insufficient attention between the conjunction words or discourse markers and the sentences.

\section{Conclusion}

We have created a benchmark test suite to evaluate the ability of pre-trained language models on seven discourse cohesion phenomena. And we consider the cohesion phenomena between adjacent sentences/clauses and non-adjacent sentences. Moreover, we conduct analysis on the results of different pre-trained language models for six discourse cohesion phenomena. In the future, we would like to know the capability of language models in terms of global cohesion.

\section*{Acknowledgments}
We would like to thank the anonymous reviewers for their insightful comments. The corresponding author is Deyi Xiong (dyxiong@tju.edu.cn).

\normalem

\bibliography{acl_latex}

%\clearpage
\newpage

\end{document}